\documentclass[conference]{IEEEtran}
\IEEEoverridecommandlockouts
\usepackage{cite}

\usepackage{amsmath,amssymb,amsfonts}
\usepackage{graphicx}
\usepackage{textcomp}
\usepackage{xcolor}
\def\BibTeX{{\rm B\kern-.05em{\sc i\kern-.025em b}\kern-.08em
    T\kern-.1667em\lower.7ex\hbox{E}\kern-.125emX}}
\definecolor{cvprblue}{rgb}{0.21,0.49,0.74}
\usepackage[pagebackref,breaklinks,colorlinks,allcolors=cvprblue]{hyperref}
\usepackage{cleveref}
\usepackage{booktabs}

\usepackage{xspace}
\usepackage{comment}
\usepackage{adjustbox}
\usepackage{array}
\usepackage{makecell}
\usepackage{pifont}
\usepackage{gensymb}
\usepackage{multirow}
\usepackage{flushend}
\usepackage{cuted}
\usepackage{capt-of}

\usepackage{algorithm} 
\usepackage[noend]{algpseudocode} 
\usepackage{placeins}
\usepackage{float}

\makeatletter
\DeclareRobustCommand\onedot{\futurelet\@let@token\@onedot}
\def\@onedot{\ifx\@let@token.\else.\null\fi\xspace}

\def\eg{\emph{e.g}\onedot} 
\def\ie{\emph{i.e}\onedot}

\makeatother

\usepackage{pifont}% http://ctan.org/pkg/pifont
\usepackage{dblfloatfix}

\newcommand{\TODO}[1]{\textbf{\color{red}[TODO: #1]}}
\renewcommand{\TODO}[1]{}

\usepackage{tikz}

\newcommand\copyrighttext{%
  \footnotesize \textcopyright \the\year{} IEEE. Personal use of this material is permitted. Permission from IEEE must be obtained for all other uses, including reprinting/republishing this material for advertising or promotional purposes, collecting new collected works for resale or redistribution to servers or lists, or reuse of any copyrighted component of this work in other works.}

\newcommand\submittedtext{%
  \footnotesize This work has been submitted to the IEEE for possible publication. Copyright may be transferred without notice, after which this version may no longer be accessible.}

\newcommand\submittednotice{%
\begin{tikzpicture}[remember picture,overlay]
\node[anchor=south,yshift=10pt] at (current page.south) {\fbox{\parbox{\dimexpr0.65\textwidth-\fboxsep-\fboxrule\relax}{\submittedtext}}};
\end{tikzpicture}%
}

\definecolor{mygray}{gray}{0.85}
\usepackage[table]{colortbl}

\newcommand{\andreic}[1]{}
\newcommand{\andrei}[1]{#1}
\newcommand{\simonc}[1]{}
\newcommand{\simon}[1]{#1}
\newcommand{\mathiasc}[1]{}
\newcommand{\mathias}[1]{#1}
\newcommand{\hassanc}[1]{}
\newcommand{\hassan}[1]{#1}

\begin{document}

\title{DOC-Depth: A novel approach for dense depth ground truth generation}

\author{\IEEEauthorblockN{de Moreau Simon}
\IEEEauthorblockA{Mines Paris -- PSL \& Valeo, France \\
simon.de\_moreau@minesparis.psl.eu}
\and
\IEEEauthorblockN{Corsia Mathias}
\IEEEauthorblockA{Exwayz, France \\
mathias.corsia@exwayz.fr}
\and
\IEEEauthorblockN{Bouchiba Hassan}
\IEEEauthorblockA{Exwayz, France \\
hassan.bouchiba@exwayz.fr}
%\and

\and[\hfill\mbox{}\par\mbox{}\hfill]
\IEEEauthorblockN{Almehio Yasser}
\IEEEauthorblockA{Valeo, France \\
yasser.almehio@valeo.com}
\and
\IEEEauthorblockN{Bursuc Andrei}
\IEEEauthorblockA{Valeo AI, France \\
andrei.bursuc@valeo.com}
\and
\IEEEauthorblockN{El-Idrissi Hafid}
\IEEEauthorblockA{Valeo, France \\
hafid.el-idrissi@valeo.com}
\and
\IEEEauthorblockN{Moutarde Fabien}
\IEEEauthorblockA{Mines Paris -- PSL, France \\
fabien.moutarde@minesparis.psl.eu}
}

\maketitle
\submittednotice

\begin{strip}
    
    \centering
    
    \vspace{-40pt}
    \includegraphics[width=0.88\textwidth]{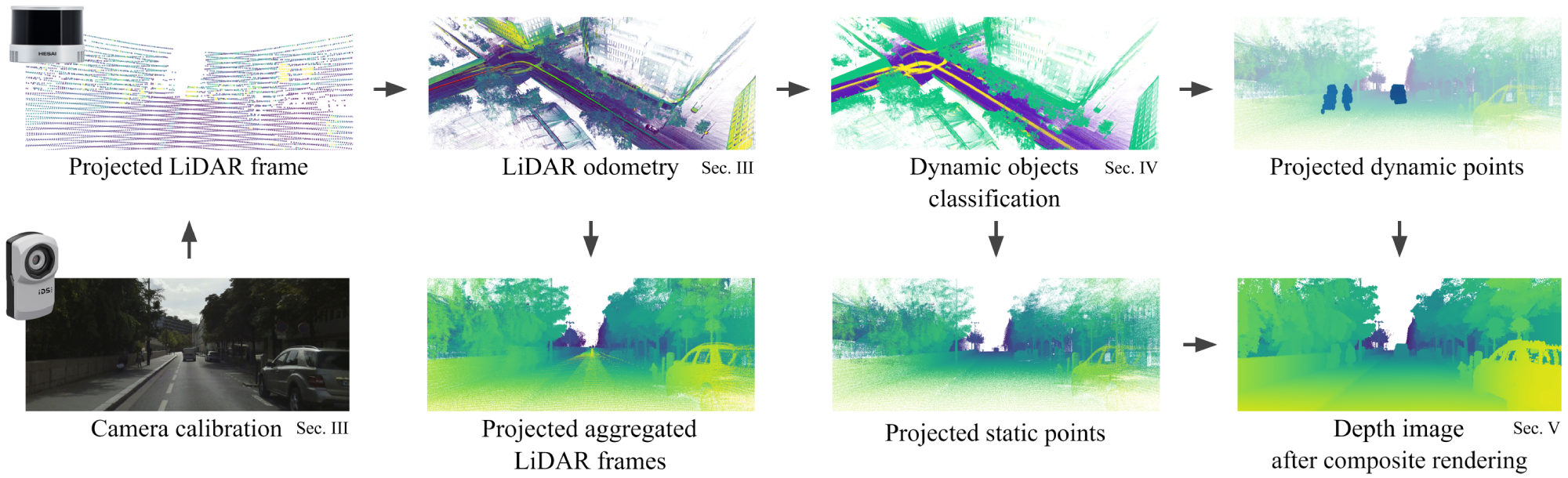}
    \vspace{-5pt}
    \captionof{figure}{
    \textbf{DOC-Depth generates dense and accurate depth ground truth \andrei{for training camera-based depth estimation systems.}} First, we aggregate LiDAR frames to obtain a 3D dense representation of the scene. Then, thanks to DOC, we classify dynamic points to handle them with specific rendering. Finally, we project the 3D reconstruction into the camera point of view, taking into account points distance and dynamic objects occlusions. 
    }
        \vspace{-10pt}

    \label{fig:pipeline}
\end{strip}

\begin{abstract}

Accurate depth information is essential for many computer vision applications. Yet, no available dataset recording method allows for fully dense accurate depth estimation in a large scale dynamic environment. In this paper, we introduce DOC-Depth,  a novel, efficient and easy-to-deploy approach for dense depth generation from any LiDAR sensor. 
After reconstructing consistent dense 3D environment using LiDAR odometry, we address dynamic objects occlusions automatically thanks to DOC, our state-of-the art dynamic object classification method. Additionally, DOC-Depth is fast and scalable, allowing for the creation of unbounded datasets in terms of size and time. We demonstrate the effectiveness of our approach on the KITTI dataset, improving its density from 16.1\% to 71.2\% and release this new fully dense depth annotation, to facilitate future research in the domain. We also showcase results using various LiDAR sensors and in multiple environments. All software components are publicly available for the research community at \href{https://simondemoreau.github.io/DOC-Depth/}{https://simondemoreau.github.io/DOC-Depth/}.
\end{abstract}

\section{Introduction}

\begin{figure}[t]
    \centering
    \includegraphics[width=\columnwidth]{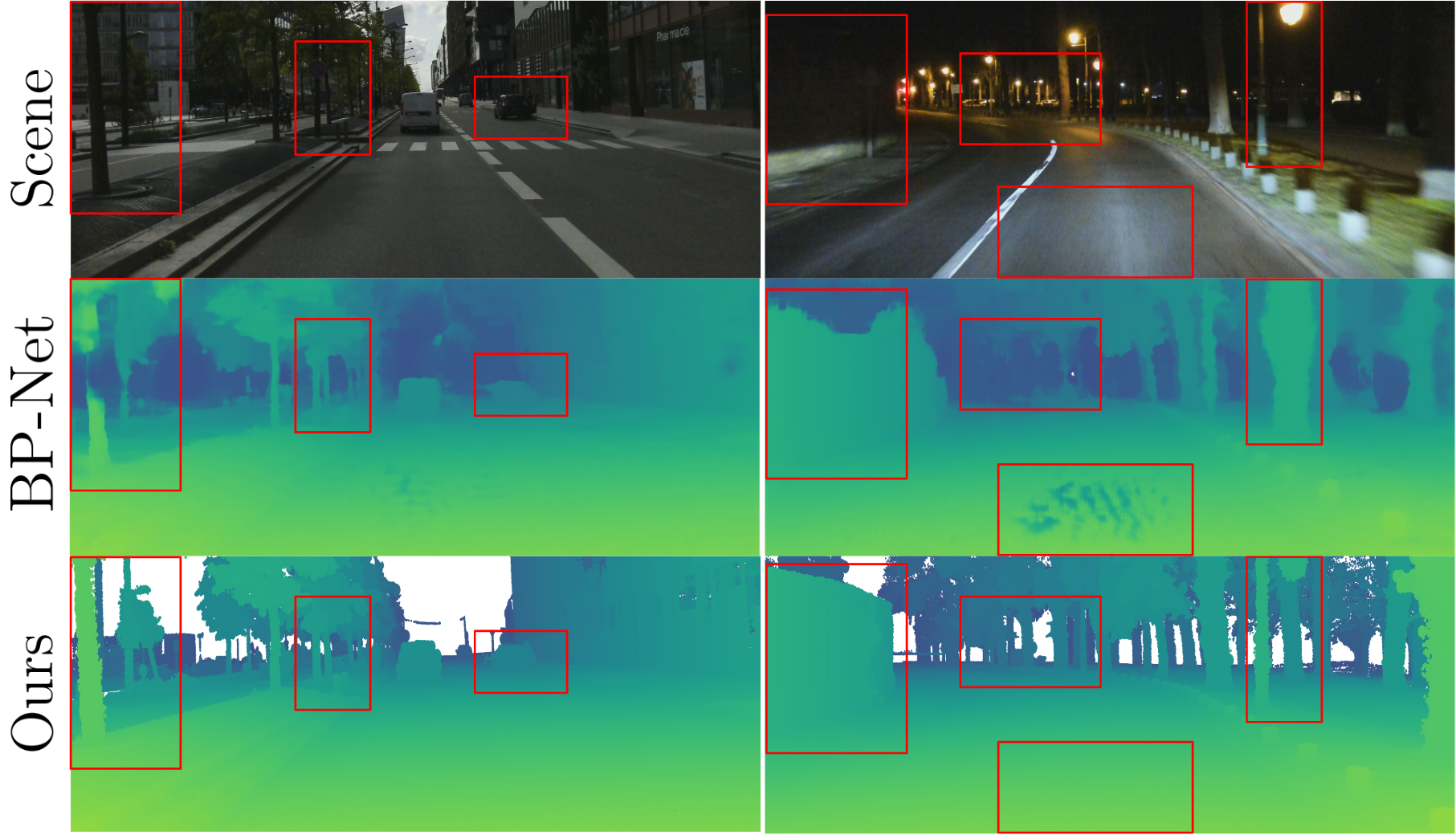}
    \vspace{-15pt}
    \caption{\textbf{Qualitative results of DOC-Depth against the learning-based approach BP-Net} \cite{tang2024bilateral} trained on KITTI and tested on our datasets. While learning-method performances drop when tested out of its training domain, our method works across domains, with the same parameters. LiDAR used: 2x Hesai XT-32 (left) and Ouster OS1-128 (right)}
    \label{fig:related_work_fail}
    \vspace{-15pt}
\end{figure}

For decades, depth information has been a crucial feature in most datasets %. 
 due to its central role in robotics and autonomous driving applications. 
Such annotations are typically acquired using sensors with limitations. While RGB-D cameras are efficient indoors, their performance degrades in sunlight \cite{brahmanage_outdoor_2019}. 
\simon{LiDAR-based solutions capture long-range depth measurements outdoors, day and night, but they provide sparse data. To address the lack of high-quality annotations, monocular depth estimation often rely on self-supervised learning methods based solely on images. While effective, these methods are less robust in adverse weather conditions \cite{gasperini_morbitzer2023md4all}. Additionally, they suffer from scale ambiguity and only offer relative depth, unlike fully-supervised models that estimate absolute metric depth. \andrei{Hu et al.} \cite{hu2019revisiting} show that losses based on normals and gradients improve depth quality and edge fidelity, but these methods require fully dense depth data, limiting their use to indoor scenarios.}
%LiDAR-based solutions can capture long-range depth measurements outdoors, from day to night, but they provide only sparse information. 
%\simon{ To overcome the lack of high quality annotations, downstream task as monocular to depth estimation have widely adopt self-supervised learning method based on image only. While these methods have proven efficient, they brittle in adverse weather conditions \cite{wang2021regularizing}. Moreover, model trained with this type of supervision usually suffer from scale-inconsistency and can only provide relative depth, while a fully-supervised one can estimate absolute metric depth. \cite{hu2019revisiting} shown that losses based on normals and gradient improve the overall depth quality and edge fidelity, but such supervision can only be applied on fully dense-depth, thus they have been used only for indoor scenario.  }
To the best of our knowledge, no available dataset recording method offers precise, fully dense depth annotations for real dynamic scenes. Several works \cite{tang2024bilateral,cao2024pasco,nunes2024cvpr} have proposed learning-based approaches for sparse-to-dense 2D/3D completion, \simon{producing dense depth from a single LiDAR scan and camera image}. However, these deep learning-based methods suffer from sensor and environment shift \simon{(see \cref{fig:related_work_fail})}, making them unsuitable for creating new datasets. %\simonc{Add paragraph about context and task definition (key terms ?)}

In this paper, we introduce DOC-Depth, a novel method to produce dense depth annotations from LiDAR measurements only. Our method truly addresses large scale ground truth generation in dynamic environments. 
Our contributions can be summarized as follows: 
\begin{itemize}
    \item Our method produces high-quality, fully-dense depth. It generalizes well among various LiDARs and environments, thanks to a learning-free, geometry-based and LiDAR agnostic approach.
    \item We propose DOC, a novel, fast and scalable dynamic object classification method that outperforms state-of-the-art methods.
    \item We release a new fully-dense annotation for KITTI depth completion \cite{uhrig2017sparsity} and odometry datasets \cite{Geiger2012CVPR}. 
    \andrei{Our approach is  easy to deploy and compatible even with low-cost low-resolution LiDARs. This opens the path towards scalable generation of large depth estimation datasets.}
    \item \andrei{We make publicly available all software components.} 

\end{itemize}
This paper is structured as follows: \cref{sec:preprocessing} describes sensor calibration and odometry estimation steps. \Cref{sec:DOC_method} introduces the DOC 
% methodology. 
approach. \Cref{sec:depth_render_method} presents our composite rendering method. Finally, \cref{sec:results} reports our results.

\section{Related work}

\subsection{Depth completion}

To perform depth completion from LiDAR, unguided \cite{uhrig2017sparsity,eldesokey2020uncertainty,eldesokey2019propagating} and image-guided \cite{zhang2023completionformer,nazir2022semattnet,tang2024bilateral} methods have been proposed. Unguided methods usually lack structural information about the scene due to the high sparsity of the input, leading to output without proper geometry. Image-guided methods use multi-branch networks and spatial propagation networks to merge image and LiDAR modalities. They produce better results with respect to the scene geometry. However, their performance can be significantly affected by adverse conditions (night, rain, etc.), that considerably damage the perception of the camera \cite{SDV21}.

While providing great performances, deep-learning methods do not generalize well on new domains \simon{as shown in \cref{fig:related_work_fail}}. Unguided methods depend on the LiDAR-pattern and image-guided ones are subject to image domain shift. Thus, they are not suited to dataset creation.  

Our method is learning-free and based only on LiDAR measurements. It achieves high performances across domains and is agnostic to LiDAR scan pattern and resolution.

\subsection{3D scene completion}
3D scene completion holds great potential for acquiring dense information.
Several approaches have been proposed to construct consistent dense scenes from sparse LiDAR measurements. 
\cite{vizzo2022make,dai2020sg,weder2021neuralfusion,murez2020atlas} are based on volumetric fusion. More recently, \cite{cao2024pasco,li2023voxformer,xia2023scpnet} propose combining semantic information to enhance results. \cite{nunes2024cvpr} successfully applied diffusion models to large scale 3D scenes. Since the input data are not sufficient to resolve all ambiguities, these methods hallucinate part of missing geometry \cite{roldao20223d}. While being interesting for mapping and understanding scenes, this is not sufficient for ground truth generation. 

DOC-Depth uses only real measurements from the LiDAR sensors, ensuring an accurate and true to life geometry.

\subsection{Dense depth datasets}
KITTI dataset \cite{Geiger2012CVPR} is a widely used benchmark for depth estimation and completion \cite{uhrig2017sparsity}. While providing large scale data, it only features semi-dense depth. Depth images are produced by aggregating 11 Velodyne HDL-64E scans and comparing them to depth values obtained through a semi-global-matching method (SGM). More recently, DDAD \cite{packnet} provided semi-dense depth annotation with only one frame from high-resolution Luminar-H2 sensors. 

ETH3D \cite{Schops_2019_CVPR} and DIODE \cite{diode_dataset} feature real dense depth annotations with measurements based on a FARO scanner. This sensor is highly accurate but requires several minutes to capture a single static scan and hours to record an entire scene, making it unsuitable for dynamic environments.

In contrast, DOC-Depth produces accurate fully-dense depth maps using any LiDAR. It results in an easier and cost-effective method for dataset creation.

\section{Pre-processing}
\label{sec:preprocessing}

DOC-Depth generates dense depth ground truth from LiDAR points projected into the camera point of view. We assume that our sensor setup contains at least one LiDAR. \andrei{We emphasize that} \simon{the camera is exclusively used for the downstream task and is not necessary to our method.}
\Cref{fig:pipeline} shows the method pipeline. 

\subsection{Calibration}

To align LiDAR and RGB camera, we compute both camera intrinsic and extrinsic calibration in the LiDAR reference frame $C^{c \rightarrow l} = \big(\begin{smallmatrix}
R^{c \rightarrow l} & u^{c \rightarrow l} \\
0 & 1 \\
\end{smallmatrix}\big) \in SE(3)$.

\andrei{Most approaches for such calibrations are either target-based \cite{zhang2000flexible} or target-free \cite{koide2023general}.}
Since we use a limited number of sensors, we opt for a target-based solution. It is usually more reliable, although more time consuming. 
Yet, we improve time efficiency by estimating both intrinsic and extrinsic calibration with the same target in a single calibration process.

We capture synchronized LiDAR-camera frames of a checkerboard target \cite{zhang2000flexible}.
We compute the distortion parameters and intrinsic matrix $K$ using \cite{zhang2000flexible}.
Subsequently, we work with rectified images, where $K$ is the rectified intrinsic matrix.
Additionally, \cite{zhang2000flexible} provides the plane parameters for each checkerboard image, expressed in the camera reference frame $\mathcal{P}^c=\{n^c, d^c, p^c\}$, with $p^c$ the origin of the plane and $x$ belongs to $\mathcal{P}^c$ if it satisfies $x . n^c - d^c = 0$.

Then, we extract each checkerboard plane in the corresponding LiDAR point cloud using RANSAC~\cite{schnabel2007efficient}: $\mathcal{P}^l=\{n^l, d^l\}$. We ensure that the normal vectors $n^c$ and $n^l$ are oriented towards their respective sensors. 

At this stage, we have a set of matching planes $\{ \mathcal{P}_i^l, \mathcal{P}_i^c \}$. We obtain $R^{c \rightarrow l}$, by minimizing \cref{eq:calib_rotation} using SVD decomposition. The translation part $u_{c \rightarrow l}$ is then computed by solving the linear system \cref{eq:calib_translation}. 
\begin{gather}
    R^{c \rightarrow l} = \underset{R \in SO(3)}{\mathrm{argmin}} \sum_i{ \lVert n_i^l - R \, n_i^c \rVert }
    \label{eq:calib_rotation} \\
    u^{c \rightarrow l} = \underset{u \in \mathbb{R}^3}{\mathrm{argmin}} \sum_i{ \lVert n_i^l . \left( R^{c \rightarrow l} \, p_i^c + u \right) - d_i^l \rVert }
    \label{eq:calib_translation}
\end{gather}

The checkerboard size should allow for plane segmentation within the point cloud.
We opt for an A0 checkerboard to ensure compatibility with any LiDAR resolution. A larger checkerboard also reduces the error on the estimated normal vector of the plane.

\subsection{Aggregating LiDAR frames}

\andrei{A key step for obtaining dense depth is to effectively merge consecutive LiDAR frames. }
To this end, we estimate both the relative motion between consecutive LiDAR frames and the intra-frame distortion, so called ego-motion.

We use Exwayz LiDAR SLAM \cite{exwayz_nav} as odometry estimator.
Freely available for academic purposes, it features low drift and is able to correct ego-motion distortion only using a LiDAR sensor. Compared to LiDAR-inertial methods \cite{xu2022fast}, it does not require extra IMU sensor to achieve low drift odometry, simplifying the setup.

Although LiDAR odometry methods tend to suffer from drift, our reconstruction only needs to be accurate over a distance equal to the maximal depth, making it compatible with any low-drift methods \cite{vizzo2023kiss,dellenbach2022ct}. 

At this stage, we have a set of undistorted LiDAR frames $\mathcal{F}_i$ and their poses $T_i \in SE(3)$ in an arbitrary local reference frame (\eg, the first frame).

\section{DOC: Dynamic Object Classification}
\label{sec:DOC_method}
\andrei{By aggregating LiDAR frames we obtain a dense and sharp point cloud, however dynamic objects still introduce unwanted geometric artifacts.}
"Dynamic trails" cause occlusions and inaccuracies in the projected depth.
We classify them for dedicated rendering.

\cite{lim2021erasor, zhang2024erasor++} present a learning-free method based on a frame-by-frame ground segmentation followed by a scan test ratio on remaining points. To perform the \textit{dynamic} - \textit{static} classification, \cite{fu2022mapcleaner} conducts ground terrain modeling on a densely aggregated point cloud, followed by a global voting method. Despite the impressive detection rate, the cleaned point clouds present \andrei{critical defects for} depth rendering. 
\simon{\cite{jia2024beautymap} uses a binary-encoded matrix as a map. It outperforms state-of-the-art methods and removes dynamic points efficiently. However, its performance depends on predefined cell sizes, limiting adaptability across environments.}

We propose a scalable, fast, and accurate approach to perform point-wise classification of each LiDAR frame into \textit{static} and \textit{dynamic} classes. 
Our novel ground segmentation method is better suited to dynamic object classification. Additionally, we propose a new voting scheme relying on sparsely sampled key frames, leveraging the ground segmentation to improve accuracy on the bottom of objects.

\begin{figure}[t]
    \centering
    \includegraphics[width=0.88\linewidth]{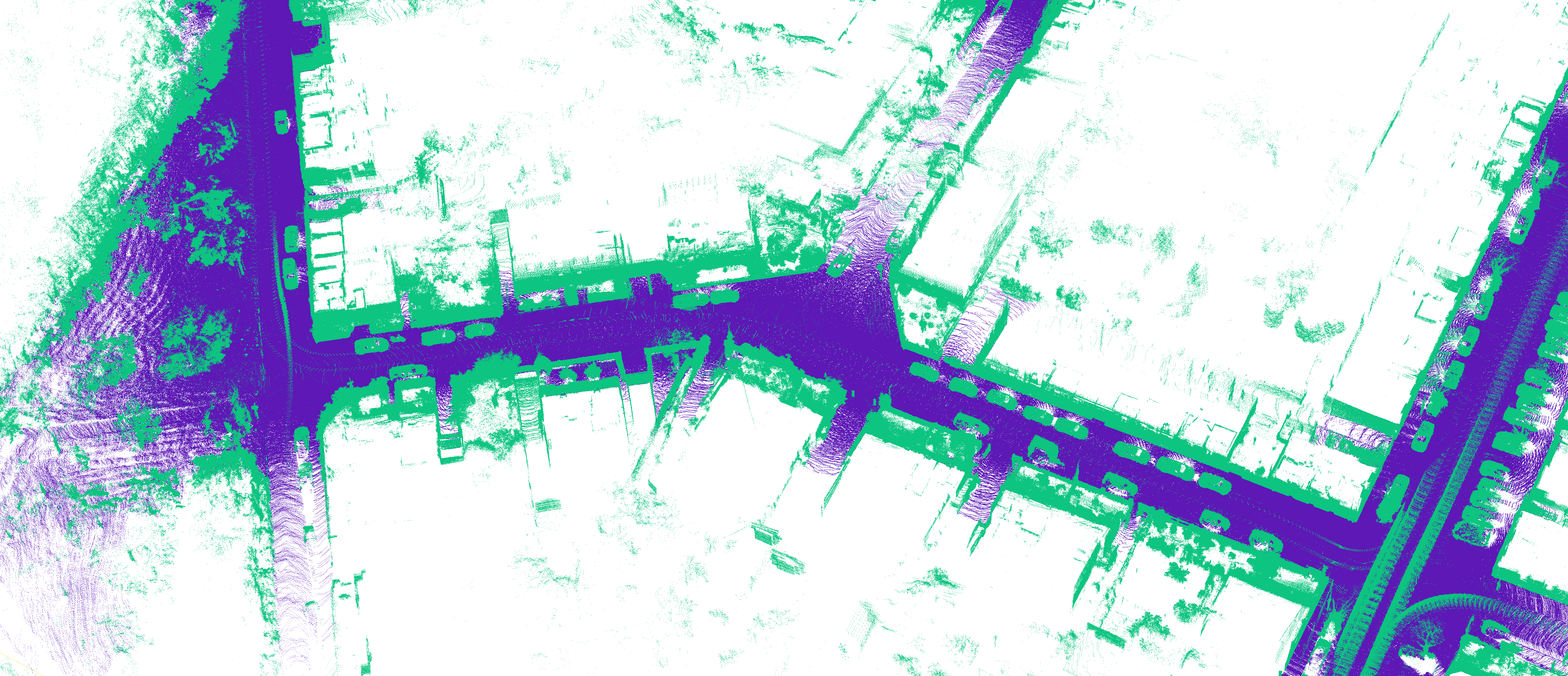}
\caption{\textbf{Example of ground segmentation results} on KITTI odometry sequence 07. 
Ground points (purple) and non-ground points (green) are accurately classified.}
\label{fig:ground_seg}
\vspace{-15pt}
\end{figure}

\subsection{Ground segmentation}
This step speeds-up the voting step by removing ground points, which are static by nature. It also enhances the classification accuracy on the bottom of dynamic objects.

Since the LiDAR sequence can be arbitrary long, we split the trajectory into chunks of length $l$.
Beyond making the approach usable on datasets of any length, this also gets rid of drift deformation which could induce non globally flat areas on long trajectory. $L$ denotes a subset of the sequence on which we aim to classify ground points. 

The frames of $L$ are merged into a single dense point cloud $\mathcal{M}_L$ by using the LiDAR odometry trajectory.
First, we find seeds on ground-likely points, leveraging the poses translation $(u_x^L \  u_y^L \ u_z^L)^T$. 
To this end, we search all the neighbors of each $(u_x^L \ u_y^L)^T$ from $\mathcal{M}_L$ in the $(X,Y)$ plan, within a radius $r_{seed}$ and we keep as seed the point with the lowest $z$ value. 
Assuming the sensor $z$ axis points upwards, the methodology proves to be accurate and robust against dynamic trails in $\mathcal{M}_L$.

We propagate \andrei{the seeds} on their $k_{nn}$ nearest neighbors using a point-to-plane distance threshold $\delta$. 
We filter points based on the verticality of their normal: points whose normal angle to the sensor $Z$ is greater than $\alpha$ are rejected from the propagation front.
Points belonging to the region grown by this process are labeled as ground points. An example of ground classification is shown in \Cref{fig:ground_seg}. 

For speed-up purposes, the input point cloud is priorly downsampled at a resolution, $s$, with a voxel-based method and labels are reprojected on the full point cloud. Unlike most ground detection methods that rely on 2D representations, our approach processes the 3D point cloud directly, enabling sharper ground segmentation without relying on a plane ground model.

\subsection{Key frame selection}
\label{sec:kf_selection}
We define a key frame as a %frame
point cloud selected to vote for others in the sequence. This step consists in selecting neighbors $\{(\mathcal{F}_j,T_j)\}_{j\in J}$ of the query frame, $(\mathcal{F}_i, T_i)$, to perform the voting operation. 
\andrei{For efficiency, we propose to select key frames sparsely through the spatial sampling of the odometry trajectory.}

We compute two subsampled versions of the odometry trajectory, the first, $\mathcal{T}_{fine}$, is finely sampled with a minimum distance of $s_{fine}$ between consecutive poses, and the second, $\mathcal{T}_{coarse}$ with a larger distance $s_{coarse}$.

\begin{figure}[t]
    \centering
    \includegraphics[width=0.9\linewidth]{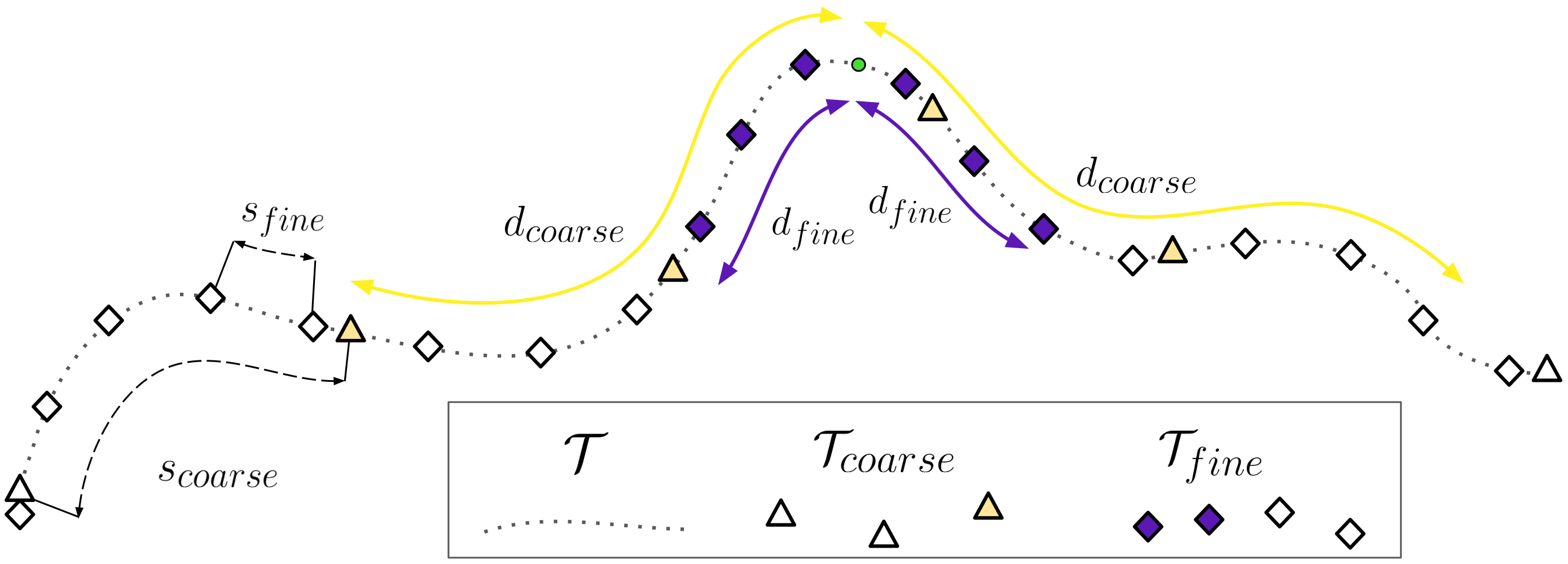}
\caption{\textbf{Illustration of the key frame selection method} on a query frame (green dot).
Colored triangles (respectively squares) correspond to the selected key frame in the coarsely (respectively finely) subsampled trajectory.}
\label{fig:kf_selection}
\vspace{-15pt}
\end{figure}

Considering the query frame $(\mathcal{F}_i, T_i)$, its adjacent key frames set $J$ is composed of all point clouds $\mathcal{F}_j$ with a pose $T_j$ in $\mathcal{T}_{coarse}$ (respectively $\mathcal{T}_{fine}$) such that $\left|d(T_j, T_i)\right| < d_{coarse}$ (respectively $d_{fine}$), where $d(T_i, T_j)$ is the signed curvilinear distance between $T_i$ and $T_j$, as depicted in \cref{fig:kf_selection}.

The idea is to finely sample key frames around the query frame $\mathcal{F}_i$ within a short distance $d_{fine}$ for accurate voting on points near the sensor, while the coarse selection, sampled over a larger distance $d_{coarse}$, handles farther objects. 

\subsection{Voting method}

We turn each key frame $(\mathcal{F}_j, T_j)$ into a spherical projection image (\ie, range image) $I_{j}$ with vertical and horizontal angular resolutions $d\phi$ and $d\theta$.
For this purpose, we discretize the angular components of the spherical coordinates $(\theta_k, \phi_k )$ of each $p_k=(x_k \ y_k \ z_k)^T \in \mathcal{F}_j$ expressed as:
\begin{equation}
\begin{aligned}
    \phi_k &= |\arccos{(z_k / \left \lVert p_k\right \rVert)}| \\
    \theta_k &= -\arctan2(y_k, x_k)
\end{aligned}
\label{eq:spherical_projection}
\end{equation}
and insert each point inside a range image.
The value in the pixel $(r_k, c_k)$ is the minimum range $\rho_k=\left \lVert p_k\right \rVert$ of all points from $\mathcal{F}_j$ falling inside this pixel.
Then, we transform each point $q \in \mathcal{F}_i$ into the key frame reference and project them on $I_j$ using  \cref{eq:spherical_projection}. 
The voting operation consists in comparing $\rho_i$ with $I_j(u, v)$, given a tolerance $\tau$. For robustness, we compare $\rho_i$ with all pixels inside a window $W$ centered on $(r_i, c_i)$ of size $w$ in pixels. The voting scheme is described in \cref{alg:voting}.

\begin{figure}[t]
    %\vspace{-7pt}
    \begin{algorithm}[H]
    \caption{DOC Voting} 
    \begin{algorithmic}[1]
        \Require{$\tau>0$}
        \State $vote \leftarrow none$
        \State $is\_ground\_pixel \leftarrow False$
        \If {$(r_i, c_i)$ $\in$ $I_j$}
            \If {$I_j(u,v)$ is ground}
                \State $\tau \leftarrow 0.0$
                \State $is\_ground\_pixel\leftarrow True$
            \EndIf
            \For {$(r, c) \in W$}
                \If {$|\rho_i - I_j(r, c)| < \tau $}
                    \State $vote\leftarrow static$
                    \State Return $vote$
                \ElsIf {$\rho_i < I_j(r, c) - \tau$}
                    \State $vote\leftarrow dynamic$
                \ElsIf {$!is\_ground\_pixel$}
                    \State $vote\leftarrow none$
                    \State Return $vote$
                \EndIf
            \EndFor
        \EndIf
        \State Return $vote$
    \end{algorithmic}
\label{alg:voting}
\end{algorithm}
\vspace{-20pt}
\end{figure}

\begin{figure}[t]
%\vspace{-20pt}
    \centering
    \includegraphics[width=0.88\linewidth]{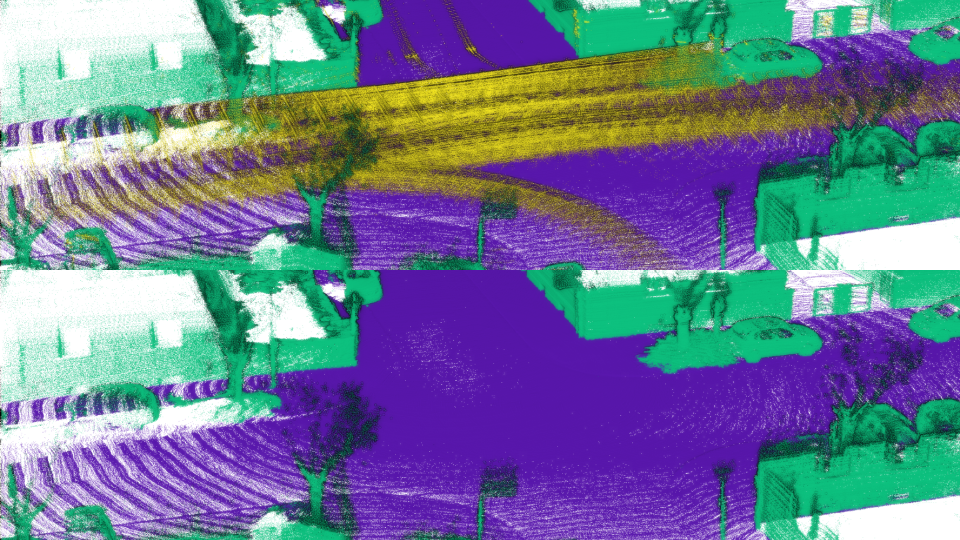}
\caption{\textbf{Example of cleaning on KITTI odometry sequence 07}. The top image shows aggregated LiDAR frames classified with DOC. The bottom image shows the quality of mobile object removal using our method. No dynamic point near the ground or floating remains.}
\label{fig:map_cleaner_single_frame}
\vspace{-15pt}
\end{figure}

The voting method is based on the idea that each LiDAR key frame defines a free space between the sensor and each measured point.
If a projected point $q'$ falls inside this free-space, it means that it belongs to a $dynamic$ object, whereas if it is close enough from the free space borders (within $\tau$), the point probably belongs to a $static$ structure.
A key contribution of our voting method is the strict handling of points causing ground occlusions by setting $\tau{=}0$. This ensures consistent voting of $dynamic$ on points belonging to the bottom of dynamic objects, e.g., car wheels or pedestrian feet.   
Given a point $q\in\mathcal{F}_i$, $c_S$ and $c_D$ values respectively denote the number of votes for $static$ and $dynamic$ classes. 
The point is finally labeled as $dynamic$ if $c_D > c_S$. 
We illustrate a qualitative result in  \Cref{fig:map_cleaner_single_frame}.

\subsection{\simon{Parameters guidelines}}
\label{sec:DOC_param}
\simon{Since DOC parameters are generic across domains and sensors, we use the same set for all our experiments.}
\simon{The $\delta$ point-to-plane threshold for the ground detection method should match the sensor noise (e.g., $4\,\text{cm}$). An exhaustive key frame selection method would use fine sampling $d_{fine}$ at a large radius $R_{coarse}$. 
\andrei{However,} a very small $d_{fine}$ \andrei{would be computationally heavy, while adding redundant} information, thus we set $d_{fine} {=} 2\,\text{m}$. Similarly, $R_{coarse}$ should align with the sensor range's order of magnitude to avoid unnecessary comparisons. To reduce computational load and to balance the contribution of near and far frames, we set $d_{coarse} {=} 5 \times d_{fine}$ and $R_{coarse} {=} 2.5 \times R_{fine}$, ensuring that farther frames contribute less to the vote.}
\simon{An isotropic projection image with a resolution lower than the sensor's has shown good experimental results. We choose $d\theta {=} d\phi {=} 0.2^\circ$ for all sensors with vertical resolution ${\le}\,1.33^\circ$.}
\simon{We provide the numerical values of the parameters used in \cref{sec:impl_details}.}

\section{Depth rendering}
\label{sec:depth_render_method}
%\vspace{-20pt}
\subsection{Camera trajectory interpolation}
\label{sec:cam_traj}
To maintain generality and simplify integration, we assume that the LiDAR and camera are not triggered simultaneously.
Thanks to our odometry, we know the LiDAR poses at each frame timestamp.
Therefore, we need to estimate its location at each camera timestamp to perform the alignment. 

Each pose is given by $T{=}(u,q)$ with $u$ the position vector and $q$ a unit quaternion representing the orientation.  
To estimate $T(t) {=} (u(t), q(t))$ the interpolated pose at timestamp $t$, we find poses $T_i$ and $T_{i+1}$ such that $t_i \leq t \leq t_{i+1}$. Then, we perform respectively a linear and spherical lineal interpolation between $T_i$ and $T_{i+1}$: 

\begin{equation}
\begin{aligned}
    u(t) &= u_i + \alpha(t) (u_{i+1} - u_i) \\
    q(t) &= q_i(q_i^{-1}q_{i+1})^{\alpha(t)}%^{(t-t_0)/(t_1-t_0)}
\end{aligned}
\label{eq:trajectory_interpolation}
\end{equation}

\noindent
with $\alpha(t)=(t-t_i)/(t_{i+1}-t_i$).
Most LiDARs \andrei{have} a 20\,Hz frame rate, so a linear approximation is sufficient, as sensor movements are not expected to undergo significant variation between frames.

\subsection{Composite rendering}
For each RGB image with timestamp $t^c$, we select the set of LiDAR frames $\{(\mathcal{F}_k,T^l_k)\}_{k\in \mathcal{D}}$, such that $-d_{min} {<} d(T^l(t^c), T^l_k) {<} d_{max}$. With $d_{min}$, the distance behind the camera and $d_{max}$ the wanted depth range.

To accelerate rendering, we only use a subset of $\mathcal{D}$, by enforcing a small minimum distance $d_{step}$ between two consecutive frames (e.g., $20\,$cm).
Then, for each LiDAR frame $\mathcal{F}_k$, we retrieve the points classified as static. Since LiDAR precision decreases with distance, we crop the point cloud and keep only points satisfying $||p^l|| < d_{crop}$. Finally, we project into camera reference frame with \cref{eq:proj_Lidar_pix}. 

\begin{equation}
\label{eq:proj_Lidar_pix}
    p^c = \left( T^{l}(t^c) \; C^{c \rightarrow l} \right)^{-1} \; T^l_k  \; p^l
\end{equation}

Normalized pixel coordinates are obtained after multiplying by $K$ and dividing by $z$. Pinhole projection spreads apart points close to the point of view and brings together far ones, inducing gaps. Therefore, projection size is set depending on point distance to the viewpoint according to \cref{eq:size_point}.
\begin{equation}
\label{eq:size_point}
    \sigma(p) = \max\left(\frac{\sigma_{max}}{\log(||p||^2)}, \sigma_{min}\right)
\end{equation}

Considering dynamic objects points, we only use the temporally closest LiDAR frame $(F_{k^*},T_{k^*})$ to $t^c$. 
Indeed, we want to capture mobile objects as they were when the RGB picture was taken.
As we can't aggregate points, we adapt $\sigma_{min}$ and $\sigma_{max}$ to fill the gaps between dynamics points according to the LiDAR resolution to obtain a fully dense depth map. 
Moreover, the depth test during projection preserves the z-order, correctly occluding farther objects with closer ones. As a result, DOC automatically resolves the see-through object problem.
\simon{Since DOC classification is point-wise, partially dynamic objects (e.g., trees) will have different point sizes for their moving and static elements.}
Given that LiDARs are often denser horizontally than vertically, we opt to project %the 
points using a vertical ellipsis shape.

We use an OpenGL implementation to leverage GPU acceleration. Programmable shaders allow us to efficiently implement our custom composite rendering.
To extract the depth map, we retrieve the z-buffer of the rendered image. Invalid pixels are the ones with maximum depth value, they usually belong to the sky.

\section{Results} \label{sec_results}
We first evaluate DOC-Depth on KITTI \cite{behley2019semantickitti,uhrig2017sparsity}, then on long sequences and highly dynamic scenarios using 3 newly recorded datasets with different LiDAR sensors.
\label{sec:results}
\subsection{Implementation details}
\label{sec:impl_details}
Our method supports both hardware and software synchronization.
To demonstrate its ease of deployment, we opt for software synchronization using ROS, which tracks sensor frame timestamps.
We use them to perform spatial alignment (see \cref{sec:cam_traj}). Processing times are evaluated on an Intel i7-13800H, 32Go RAM and Nvidia RTX 2000 Ada laptop and demonstrate DOC-Depth scalability. We used an IDS UV-36L0XC-C camera. 
\simon{DOC parameters are the same across all experiments, including KITTI and our novel datasets, demonstrating the \andrei{generalization of the parameters}. The ground classification parameters are $s {=} 0.03\,\text{m}$, $l {=} 500\,\text{m}$, $r_{seed} {=} 2\,\text{m}$, $k_{nn} {=} 30$, $\alpha {=} 15^{\circ}$, $\delta {=} 0.04\,{m}$, and the mobile object classification parameters are $d\theta {=} d\phi {=} 0.2^\circ$, $R_{coarse} {=} 50\,\text{m}$, $d_{coarse} {=} 10\,\text{m}$, $R_{fine} {=} 20\,\text{m}$, $d_{fine} {=} 2\,\text{m}$, $w {=} 5$ and $\tau = 0.2\,\text{m}$.}

\begin{figure*}
    \centering
    \includegraphics[width=0.98\linewidth]{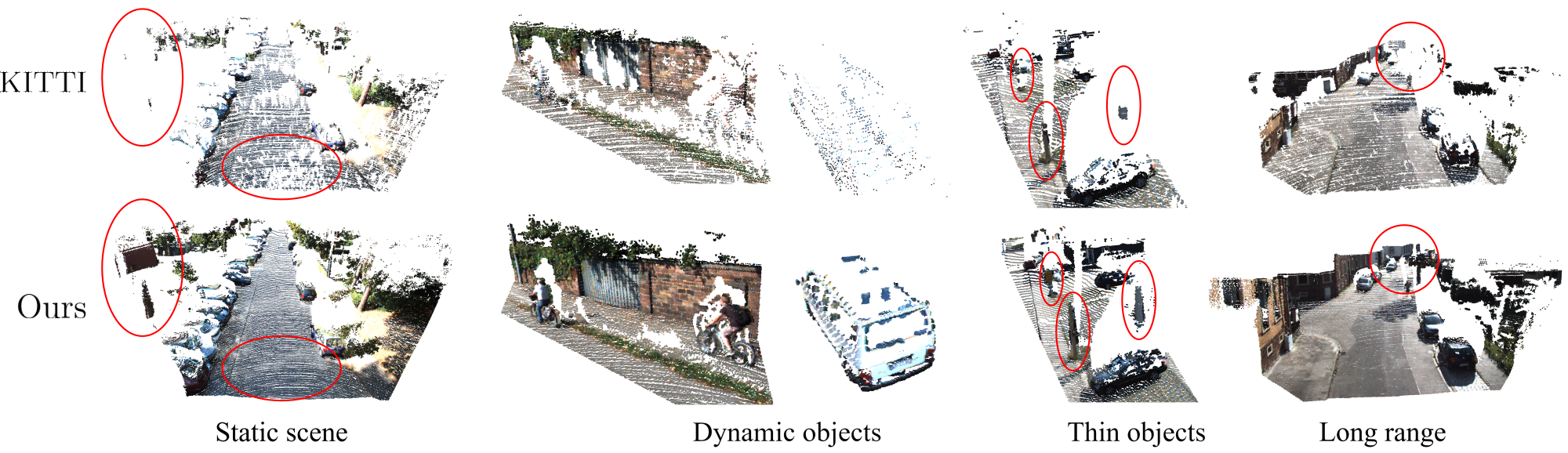}
    \caption{\textbf{Qualitative comparison between KITTI depth completion and DOC-Depth}. 
    Our method preserves all scene structures, whereas KITTI lacks untextured walls and thin objects due to SGM validation.
    KITTI’s ground is uneven with large gaps, and mobile objects are duplicated by the scan aggregation.
    In contrast, DOC-Depth ensures accurate dynamic object reconstruction and consistent geometry across the entire depth.}
    \label{fig:kitti_fail}
    \vspace{-15pt}

%\vspace{-10pt}
\end{figure*}

\subsection{Comparison on KITTI}
\subsubsection{Dynamic object classification}
\label{sec:doc_results}
\mathias{
We evaluate DOC against ERASOR \cite{lim2021erasor}, MapCleaner \cite{fu2022mapcleaner}, Dynablox \cite{schmid2023dynablox} and BeautyMap \cite{jia2024beautymap} using the Static Accuracy (SA) and Dynamic accuracy (DA) introduced in \cite{jia2024beautymap}.}
\begin{comment}
    , defined as
\begin{equation}
    SA = \frac{TP_{static}}{GT_{static}}\ ,\  DA = \frac{TP_{dynamic}}{GT_{dynamic}}\
\end{equation}
where, for a given class (static or dynamic) $TP$ represents the number of true positive and $GT$ the total number of instances.  
\end{comment}
\simon{We exclude ERASOR++ \cite{zhang2024erasor++} from our benchmark as no code is available and it reports only $PR$ and $RR$ metrics, computed on a voxel grid rather than all points. While similar to $SA$ and $DA$, these metrics are less penalizing \cite{zhang2023dynamic}.}  \hassan{We report results in \Cref{tab:comp_erasor_map_cleaner}.}

\begin{table}[tb]
\centering

\caption{Dynamic point classification results on SemanticKITTI.}
%\caption{Dynamic point classification results and comparison to ERASOR \cite{lim2021erasor}, MapCleaner, Dynablox \cite{schmid2023dynablox} and BeautyMap \cite{jia2024beautymap} on SemanticKITTI.}
    \resizebox{0.75\columnwidth}{!}{
        \begin{tabular}{ c  c  c  c  c }
        \toprule
            \textbf{Sequence} & \textbf{Method} & \textbf{SA (\%) $\uparrow$} & \textbf{DA (\%) $\uparrow$} & \textbf{F1-score $\uparrow$} \\
        \midrule
            \multirow{5}{*}{00} 
                & ERASOR & 66.70 & 98.54 & 0.7955 \\
                & MapCleaner & 98.89 & 98.18 & 0.9853 \\ 
                & Dynablox & 96.76 & 90.68 & 0.9362 \\
                & BeautyMap & 96.76 & 98.38 & 0.9756 \\
                & \cellcolor{mygray}DOC (Ours) & \cellcolor{mygray}\textbf{99.73} & \cellcolor{mygray}\textbf{98.99} & \cellcolor{mygray}\textbf{0.9935} \\ \midrule
            \multirow{5}{*}{01}
                & ERASOR & 98.12 & 90.94 & 0.9439 \\
                & MapCleaner & \textbf{99.74} & 94.98 & 0.9730 \\
                & Dynablox & 96.33 & 68.01 & 0.7373 \\
                & BeautyMap & 99.17 & 92.99 & 0.9598 \\
                & \cellcolor{mygray}DOC (Ours) & \cellcolor{mygray}99.66 & \cellcolor{mygray}\textbf{96.89} & \cellcolor{mygray}\textbf{0.9825} \\ \midrule
%            \multirow{3}{*}{02} 
%                & ERASOR & 87.73 & 97.01 & 0.921 \\
%                & MapCleaner & 99.37 & \textbf{99.03} & \textbf{0.9920} \\ 
%                & Ours & \textbf{99.73} & 98.31 &  0.9901 \\ \midrule
            \multirow{5}{*}{05} 
                & ERASOR & 69.40 & 99.06 & 0.8162 \\
                & MapCleaner & 99.14 & 97.92 & 0.9852 \\
                & Dynablox & 97.80 & 88.68 & 0.9302 \\
                & BeautyMap & 96.34 & 98.29 & 0.9731 \\
                & \cellcolor{mygray}DOC (Ours) & \cellcolor{mygray}\textbf{99.69} & \cellcolor{mygray}\textbf{99.07} & \cellcolor{mygray}\textbf{0.9937} \\ \midrule
%            \multirow{3}{*}{07} 
%                & ERASOR & 90.62 & \textbf{99.27} & 0.948 \\
%                & MapCleaner & 98.98 & 97.25 & 0.9811 \\ 
%                & Ours & \textbf{99.81} & 99.04 & \textbf{0.9942} 
                %\\ \midrule
            \multirow{5}{*}{Average} 
                & ERASOR & 78.07 & 96.18 & 0.8618 \\
                & MapCleaner & 99.25 & 97.02 & 0.9812 \\
                & Dynablox & 96.96 & 82.46 & 0.8912 \\
                & BeautyMap & 97.42 & 96.55 & 0.9698 \\
                & \cellcolor{mygray}DOC (Ours) & \cellcolor{mygray}\textbf{99.69} & \cellcolor{mygray}\textbf{98.31} & \cellcolor{mygray}\textbf{0.9899} \\    
                \bottomrule
        \end{tabular}
    }
    
    \label{tab:comp_erasor_map_cleaner}
    \vspace{-15pt}
\end{table}
DOC outperforms other methods in both $SA$ and $DA$, making it ideal for depth ground truth generation. Maximizing $DA$ prevents foreground points belonging to mobile objects from being introduced during rendering, and maximizing $SA$ avoids excessive erosion of static objects. Moreover, the position of false negatives is critical: dynamic points near LiDAR poses would cause significant occlusions. Our voting method eliminates such points, ensuring accurate and sharp depth.

\subsubsection{Ablation on DOC sampling strategy}
\simon{A key contribution of DOC is its key frame sampling strategy over the trajectory. It optimizes processing time and balances vote contributions from both near and distant key frames.
We study the impact of using separate $(R_{coarse}, d_{coarse})$ and $(R_{fine}, d_{fine})$ sampling combinations against a single sampling combination $(R,d)$ on both the processing time of the voting step and classification performance. The results are averaged across all KITTI sequences selected for evaluation.}
\simon{Our baseline is compared with results obtained using single parameter combinations $\mathcal{P}_1 {=} (50\,\text{m}, 10\,\text{m})$, $\mathcal{P}_2 {=} (20\,\text{m}, 2\,\text{m})$, and $\mathcal{P}_3 {=} (50\,\text{m}, 2\,\text{m})$. We present the results in \Cref{tab:ablation}.}

\simon{This study highlights that DOC is robust to changes in key frame sampling parameters, which primarily impact the processing time.
Thanks to our precise ground classification and new voting scheme, we outperform state-of-the-art methods, even with the $\mathcal{P}_1$ parameter set, which achieves a 50\,Hz classification rate for the voting step. As discussed in \cref{sec:DOC_param}, exhaustive key frame selection does not improve performance due to high information redundancy. This is demonstrated by $\mathcal{P}_3$, which despite having more voting key frames and longer processing time, does not achieve the best $F1$-score.
Votes from key frames too far from the current frame contribute equally to the global vote, resulting in false positives in the $dynamic$ class. DOC achieves higher precision for both $static$ and $dynamic$ classes by better balancing votes from nearby and distant frames.}

\begin{table}[tb]
\centering
\caption{DOC sampling strategy ablation on SemanticKITTI.}
    \resizebox{0.8\columnwidth}{!}{
        \begin{tabular}{ c  c  c  c  c }
        \toprule
            \textbf{Parameters set} & \textbf{SA (\%) $\uparrow$} & \textbf{DA (\%) $\uparrow$} & \textbf{F1-score$\uparrow$} & \textbf{Processing time (s) $\downarrow$} \\
        \midrule 
                $\mathcal{P}_1$ & 99.54 & 97.97 & 0.9874 & \textbf{0.019}\\ 
                $\mathcal{P}_2$ & \textbf{99.72} &  98.02 & 0.9886 & 0.037 \\ 
                $\mathcal{P}_3$ & 99.63 & \textbf{98.31} & 0.9891 & 0.070\\
                \rowcolor{mygray} DOC (Ours) & 99.68 & \textbf{98.31} & \textbf{0.9899} & 0.049 \\
            \bottomrule
        \end{tabular}
    }
    \label{tab:ablation}
    \vspace{-15pt}
    
\end{table}

\begin{table}[b]
    \vspace{-20pt}

    \caption{Evalutation on KITTI depth completion.}
    \centering
    \resizebox{0.75\columnwidth}{!}{
    \begin{tabular}{c|c c c}
    \toprule
         \textbf{Point-to-Point (m) $\downarrow$}  & \textbf{RMSE (m) $\downarrow$} & \textbf{MAE (m) $\downarrow$} & \textbf{Abs Rel $\downarrow$} \\ \midrule
          0.081 & 2.07 & 0.63 & 0.033 \\
         \bottomrule
    \end{tabular}
    }
    \label{tab:kitti_res}
    
\end{table}

\begin{figure*}
    \centering
    \includegraphics[width=0.9\linewidth]{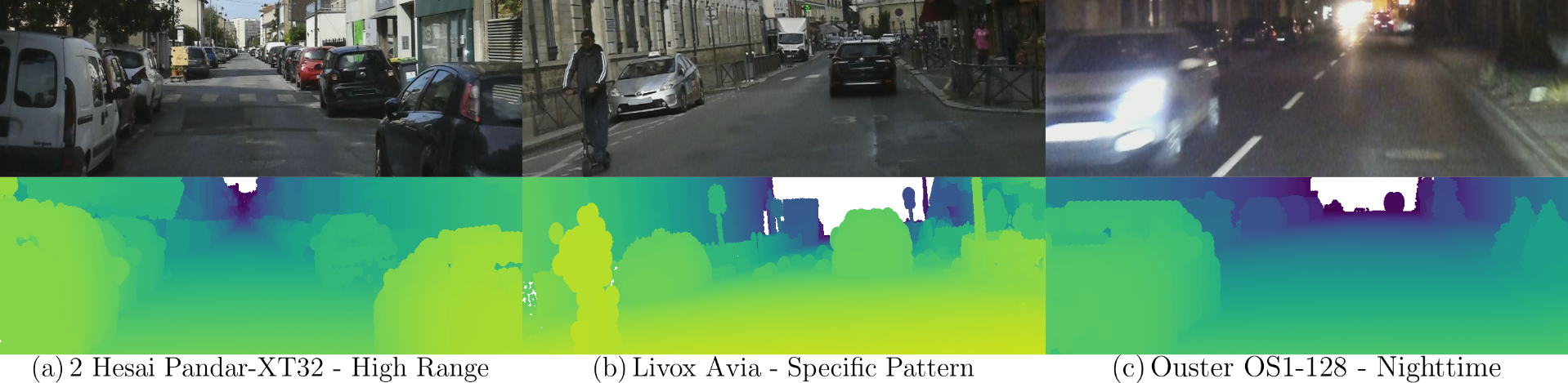}
    
    \vspace{-5pt}
    \caption{\textbf{Qualitative result using various sensors}. Our LiDAR-agnostic method produces accurate, dense depth regardless of the LiDAR pattern. By relying solely on LiDAR, ground truth can be generated from day to night.}
    \label{fig:quali_depth}
    \vspace{-15pt}
\end{figure*}
\subsubsection{Dense depth generation}
\label{sec:dense_kitti}
KITTI depth completion 
\cite{uhrig2017sparsity} is widely used as a semi-dense depth dataset (16.1\% of pixels). While 2D depth maps seem satisfying, significant failures arise when re-projecting into 3D space, as shown in \Cref{fig:kitti_fail} (first row). Due to the aggregation process, mobile objects are duplicated. The depth maps are validated using SGM, which struggles in texture-less areas and with thin vertical objects, leading to missing walls, poles, and street lamps.
In comparison, our depth is much denser (71.2\% of pixels), with mobile objects accurately positioned and all scene structures preserved.
We provide qualitative results in \Cref{fig:kitti_fail}.

To validate that our geometry is consistent with theirs, we compare the 3D deprojected depth maps using a point-to-point metric. 
Results are presented in \Cref{tab:kitti_res}. 
Our average distance of $8.14\,\text{cm}$ is approximately equivalent to the sensor noise, demonstrating that both geometries are consistent and that DOC-Depth correctly reconstructs the scenes.

While our reconstruction is highly precise, errors are observed in the 2D quantitative results.
One reason is the missing foreground objects in the KITTI depth maps, which causes background objects to be incorrectly treated as "ground truth points" (as shown in \Cref{fig:kitti_fail}). 
Also, the low quality of the KITTI INS trajectory \cite{behley2019semantickitti} used for frame aggregation causes misalignment noise and incorrect ego-motion compensation.

We will release a new fully dense version of the KITTI depth completion and odometry datasets annotations, enabling new explorations and further research in the field.

\subsubsection{Comparison to learning-based methods}
\simon{While learning-based sparse-to-dense depth completion is a related work, we cannot directly compare performances. As shown in \cref{sec:dense_kitti}, the KITTI ground truth used to train these methods is unreliable and cannot serve as a benchmark. Therefore, we propose a dedicated experiment to validate our method (see \cref{sec:novel_data}).}

\begin{figure}[t]
    \centering
    \includegraphics[width=0.8\columnwidth]{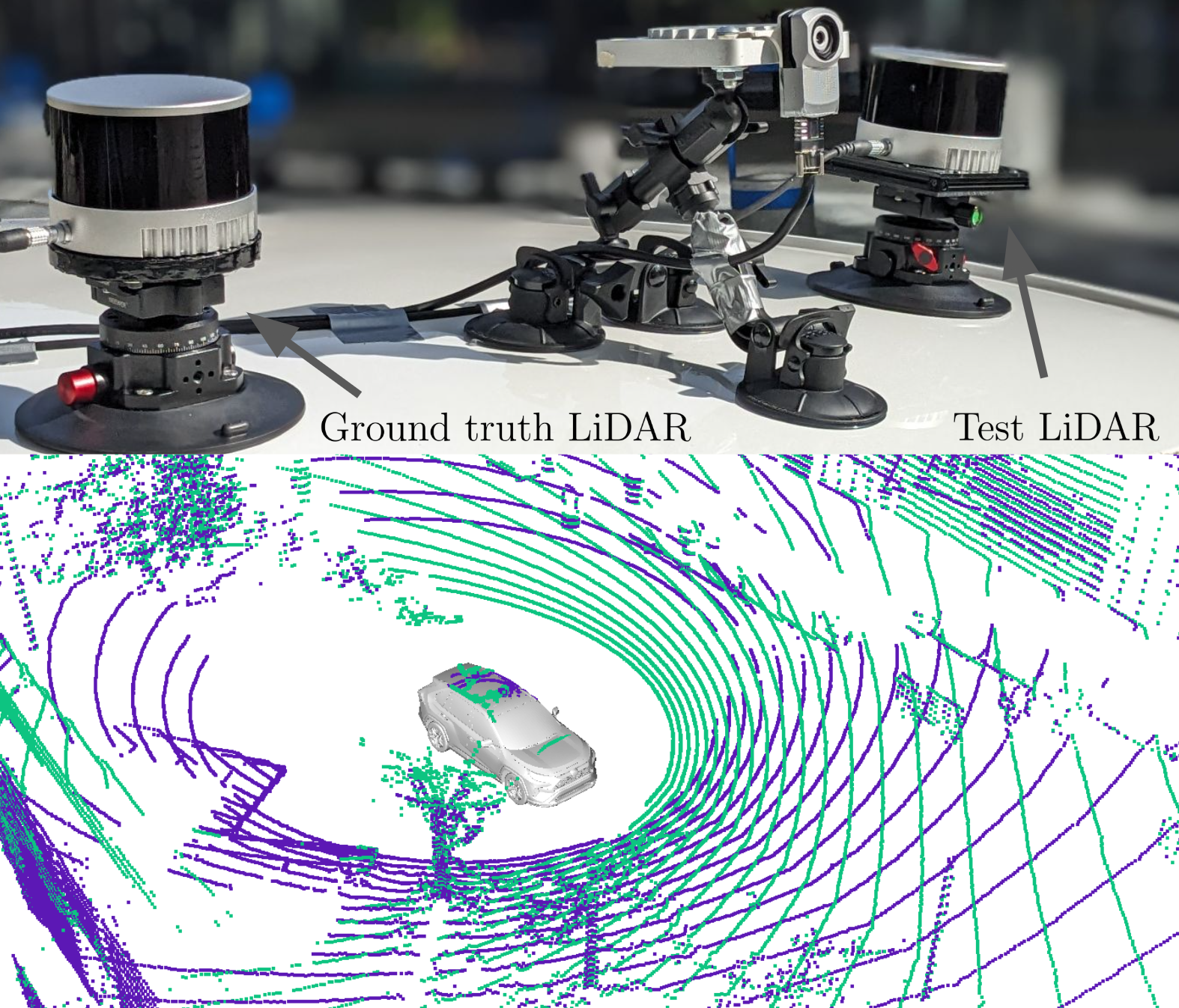}
    \caption{\textbf{Experimental sensor setup used to assess our method}. 
    The two LiDAR have different orientations with minimal overlap in raw measurements, allowing us to assess the quality of our reconstruction.}
    \label{fig:2lidar_val}
    \vspace{-15pt}
\end{figure}

\subsection{Novel datasets results}

\label{sec:novel_data}
\begin{table}[b]
% \vspace{-12.5pt}

\vspace{-20pt}
    \caption{Datasets information}
    \resizebox{\columnwidth}{!}{
    \begin{tabular}{c|c c r l c c}
    \toprule
         \textbf{LiDAR} & \textbf{Area} & \textbf{Recording Time} & \multicolumn{2}{c}{\textbf{Processing Time}} & \textbf{\# Depth Map} & \textbf{Density}  \\ \midrule
         OS1-128 & City / Country roads & 30\,min & 111\,min & {\scriptsize({\tiny$\times$}3.7)} & 13,700 & 79\% \\ \midrule 
         2 Pandar-XT32  & City / Highway & 41\,min & 128\,min  &{\scriptsize({\tiny$\times$}3.1)} & 15,500 & 76\% \\ \midrule 
         Avia  & City & 14\,min & 24\,min & {\scriptsize({\tiny$\times$}2.2)} & 3,200 & 86\% \\%\\ \midrule 
         %Pandar-XT32  & Indoor & 2\,min\,40 & 3\,min & {\scriptsize({\tiny$\times$}1.1)} & 213 & 99.3\%\\  
         \bottomrule

    \end{tabular}
    }
    \label{tab:dataset_info}
    
% \vspace{-12.5pt}
\end{table}

We show qualitative result of our datasets in \Cref{fig:quali_depth}. Static structures are sharply reconstructed, resulting in a highly detailed depth projection. 
While mobile objects are less refined, they are fully dense, ensuring correct occlusion (no see-through) and precise positioning. 
Generated depth images are almost fully dense as reported in \Cref{tab:dataset_info}.

\simon{Since existing datasets like KITTI cannot be used to assess our method's reconstruction, we designed a dedicated experiment to quantitatively validate our approach.
As shown in \Cref{fig:2lidar_val}, we created a two-LiDAR setup. We generate the depth map with only one LiDAR, reducing the density only from 76\% to 73\%. We then compare DOC-Depth output with the corresponding sparse projected frames from the other LiDAR, serving as ground truth. As each LiDAR has a different orientation, we ensure minimal overlap in the original measurements, assessing DOC-Depth reconstruction fidelity. Points affected by occlusion from sensor positions are filtered out.
Across the entire sequence, we achieve an RMSE of $0.26\,\text{m}$, demonstrating the precision of our generation method in scenarios with a high number of dynamic objects.}

\subsubsection{Method generalizability}
As DOC-Depth does not have prior on point clouds data, it is agnostic to LiDAR sensors.
\Cref{fig:quali_depth} shows depth generated using various sensors. 
Even with Livox Avia non-repetitive pattern (b), DOC-Depth produces consistent depth reconstruction.
Output quality is preserved regardless of the number of LiDAR beams (a,c).
To the best of our knowledge, we are the first LiDAR-based densification method assessed on 4 different sensors. 
\simon{All our results were obtained using the same set of parameters. It shows their generalizability across sensors and scenarios.}

\subsubsection{Downstream tasks}
\begin{figure}[t]
    \centering
    \includegraphics[width=\columnwidth]{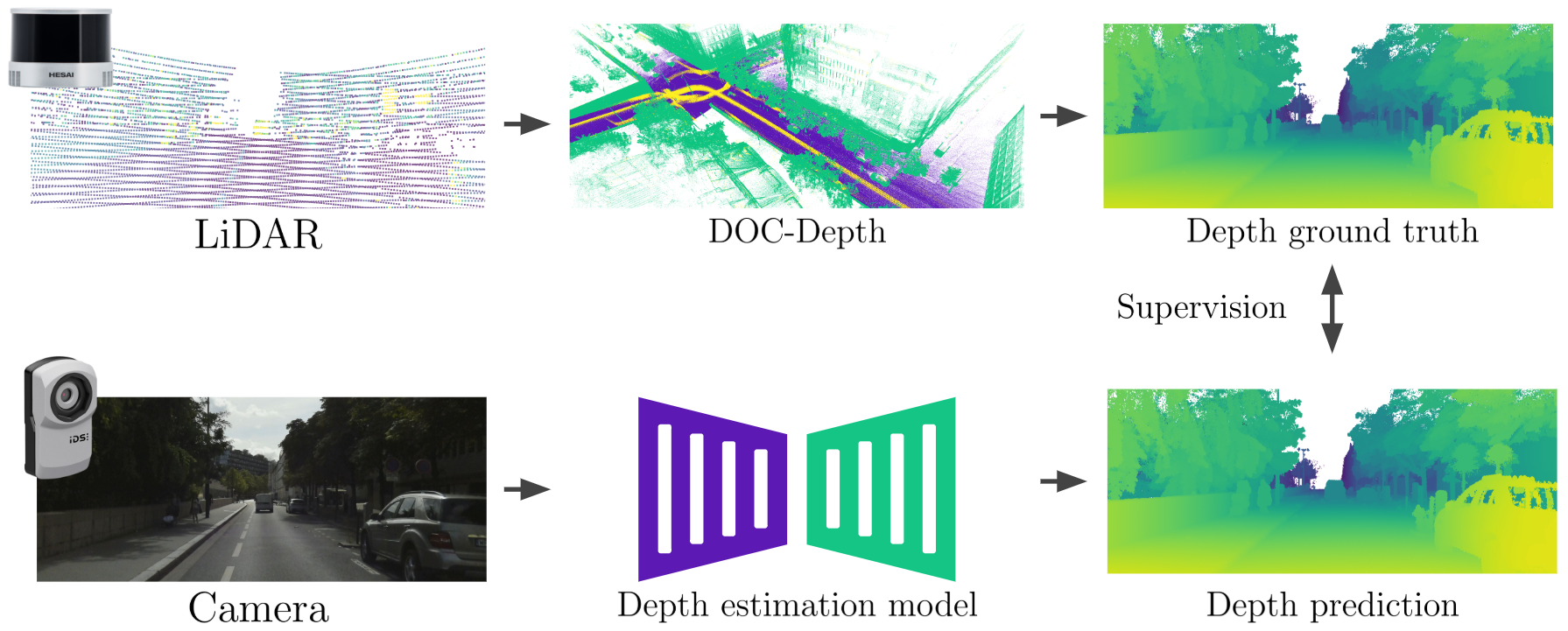}
    \vspace{-15pt}
    \caption{\textbf{Example of monocular to depth model training pipeline} as a downstream task using our method. The training code remains identical. }
    \label{fig:downstream_pipeline}
    
    % \vspace{-15pt}
    \vspace{-10pt}
\end{figure}

\simon{Using DOC-Depth for downstream tasks like monocular depth estimation is straightforward (see \cref{fig:downstream_pipeline}). Instead of projecting a single LiDAR scan into an image to supervise the model, our pipeline generates a fully dense version of the LiDAR measurements. 
\andrei{We highlight that this does not change the model training code.} With scan projections, only a few valid pixels are available for loss computation, whereas our dense data allows almost the entire image to be used for supervision, making it highly data-efficient. Since DOC-Depth relies solely on LiDAR, it performs well in low-light conditions, providing accurate ground truth at nighttime (\Cref{fig:quali_depth} (c)). Our nighttime ground truth has been successfully used to train deep learning models in \cite{demoreau2024led}. With fully dense annotations, these models benefit from \cite{hu2019revisiting} losses to improve depth quality. This demonstrates the applicability of our method to downstream tasks.}

\subsubsection{High range depth}
Thanks to the frame aggregation, we can project static points from any distance.
\Cref{fig:quali_depth} (a) shows an example of depth map with range up to $700$\,m. Only dynamic objects are limited by the sensor range. As a result, our method can reconstruct very large static scenes, even with short-range, low-cost sensors.

%\vspace{-2pt}
\section{Discussion \andrei{and limitations}}

As our geometry-based method only relies on LiDAR data, it is sensitive to incorrect measurements caused by blooming on retro-reflective objects.
They lead to unwanted noise points around them, decreasing the depth quality.

\andrei{Although DOC classifies dynamic points well, there is room for performance improvement for temporarily static objects, e.g., cars waiting at a red light, pedestrians standing still.}

Surface splatting \cite{botsch2005high} render each point as an oriented disk perpendicular to its normal. It could enhance the proposed rendering, especially on flat surfaces like the ground.
Due to the sparse nature of dynamic points, they produce the blurriest part of the depth. Integrating image segmentation \cite{kirillov2023segment} with LiDAR measurements could significantly improve the sharpness of these objects. 

%\vspace{-2pt}
\section{Conclusion}
We introduced DOC-Depth, a novel method for dense depth ground truth generation in unbounded dynamic environments. Thanks to DOC, our proposed dynamic object classifier, we automatically address objects occlusions in the image. We demonstrate the effectiveness of our approach on both KITTI and in-house captured datasets \andrei{with 4 different LiDAR types}. Through novel dataset creation using various sensors, we show its generalization and ease of deployment. We release a new fully-dense annotation for KITTI depth completion and odometry datasets, enabling further exploration in depth estimation and completion.
Software components are available for the research community. 

\bibliographystyle{IEEEtran}
\bibliography{main}

\end{document}